\documentclass[11pt,a4paper]{article}
\usepackage[hyperref]{naaclhlt2018}
\usepackage[utf8]{inputenc}
\usepackage{times}
\usepackage{latexsym}
\usepackage{graphicx}
\usepackage{amsmath,amssymb}
\usepackage{algorithm,algorithmicx}
\usepackage[noend]{algpseudocode}
\usepackage{url}
\usepackage[textsize=tiny]{todonotes}
\usepackage{xspace}
\usepackage{caption}

\aclfinalcopy 
\def\aclpaperid{362} 

\DeclareMathOperator{\argmax}{argmax}
\DeclareCaptionFont{10pt}{\fontsize{10pt}{12pt}\selectfont}
\def\undo{\ensuremath{%
  \rotatebox[origin=c]{90}{$\circlearrowleft$}}}

\newcommand{\bigO}[1]{\ensuremath{\mathcal{O}(#1)}}
\newcommand{\blackstar}{$\bigstar$}
\newcommand{\whitestar}{$\Diamond$}

\newcommand\sos{$\langle\mathrm{s}\rangle$\xspace}
\newcommand\eos{$\langle\mathrm{/s}\rangle$\xspace}
\newcommand\sockeye{{\sc Sockeye}\xspace}
\newcommand\sacrebleu{{\sc sacreBLEU}\xspace}
\newcommand\dba{{\sc DBA}\xspace}
\newcommand\gbs{{\sc GBS}\xspace}
\newcommand\cbs{{\sc CBS}\xspace}

\title{Fast Lexically Constrained Decoding with Dynamic Beam Allocation for Neural Machine Translation}

\author{Matt Post \and David Vilar\\
  Amazon Research \\ Berlin, Germany}

\date{}

\begin{document}
\maketitle
\begin{abstract}
    The end-to-end nature of neural machine translation (NMT) removes many ways of manually guiding the translation process that were available in older paradigms.
    Recent work, however, has introduced a new capability: \emph{lexically constrained} or \emph{guided} decoding, a modification to beam search that forces the inclusion of pre-specified words and phrases in the output.
    However, while theoretically sound, existing approaches have computational complexities that are either linear \cite{hokamp:2017:lexically} or exponential \cite{anderson:2017:guided} in the number of constraints.
    We present an algorithm for lexically constrained decoding with a complexity of \bigO{1} in the number of constraints.
    We demonstrate the algorithm's remarkable ability to properly place these constraints, and use it to explore the shaky relationship between model and BLEU scores.
    Our implementation is available as part of \sockeye.\footnote{\url{https://awslabs.github.io/sockeye/inference.html\#lexical-constraints}}
\end{abstract}

\section{Introduction}

One appeal of the phrase-based statistical approach to machine translation \cite{koehn:2003:statistical} was that it provided control over system output.
For example, it was relatively easy to incorporate domain-specific dictionaries, or to force a translation choice for certain words.
These kinds of interventions were useful in a range of settings, including interactive machine translation or domain adaptation.
In the new paradigm of neural machine translation (NMT), these kinds of manual interventions are much more difficult, and a lot of time has been spent investigating how to restore them (cf.\ \citet{arthur:2016:incorporating}).

\begin{figure}[t]
  \begin{center}
    \includegraphics[width=76mm]{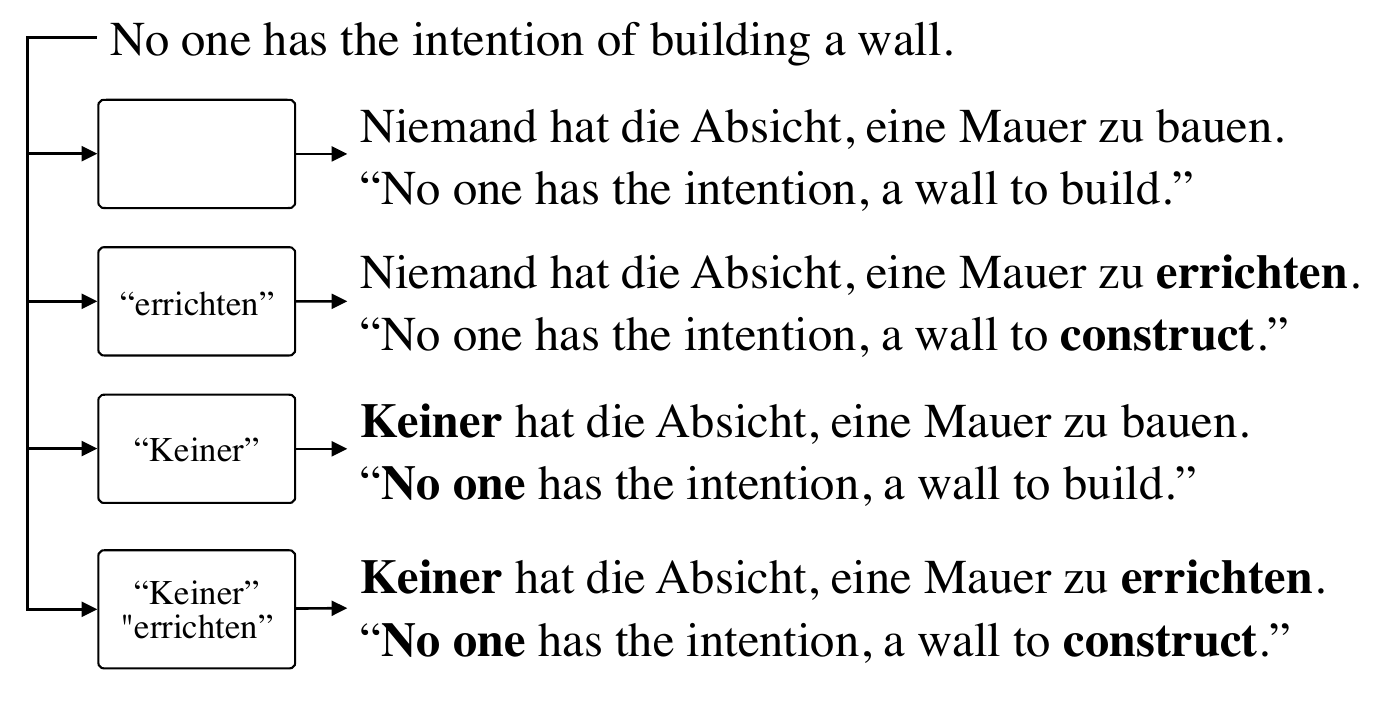}
  \end{center}
  \caption{An example translating from English to German.
  The first translation is unconstrained, whereas the remaining ones have one or two constraints imposed. A word-for-word translation of the German output has been provided for the convenience of non-German speaking readers.}
  \label{figure:example}
\end{figure}

At the same time, NMT has also provided new capabilities.
One interesting recent innovation is \emph{lexically constrained decoding}, a modification to beam search that allows the user to specify words and phrases that must appear in the system output (Figure~\ref{figure:example}).
Two algorithms have been proposed for this: \emph{grid beam search} \cite[\gbs]{hokamp:2017:lexically} and \emph{constrained beam search} \citep[\cbs]{anderson:2017:guided}.
These papers showed that these algorithms do a good job automatically placing constraints and improving results in tasks such as simulated post-editing, domain adaptation, and caption generation.

A downside to these algorithms is their runtime complexity: linear (\gbs) or exponential (\cbs) in the number of constraints.
Neither paper reported decoding speeds, but the complexities alone suggest a large penalty in runtime.
Beyond this, other factors of these approaches (a variable sized beam, finite-state machinery) change the decoding procedure such that it is difficult to integrate with other operations known to increase throughput, like batch decoding.

\begin{table}[t]
  \centering
  \begin{tabular}{l|r}
    work & complexity \\
    \hline\hline
    \citet{anderson:2017:guided} & \bigO{Nk2^C}
    \\
    \citet{hokamp:2017:lexically} & \bigO{NkC}
    \\
    This work & \bigO{Nk}
  \end{tabular}
  \caption{Complexity of decoding (sentence length $N$, beam size $k$, and constraint count $C$) with target-side constraints under various approaches.}
  \label{table:complexity}
\end{table}

We propose and evaluate a new algorithm, \emph{dynamic beam allocation} (\dba), that is \emph{constant} in the number of provided constraints (Table~\ref{table:complexity}).
Our algorithm works by grouping together hypotheses that have met the same number of constraints into \emph{banks} (similar in spirit to the grouping of hypotheses into stacks for phrase-based decoding \citep{koehn:2003:statistical}) and dynamically dividing a fixed-size beam across these banks at each time step.
As a result, the algorithm scales easily to large constraint sets that can be created when words and phrases are expanded, for example, by sub-word processing such as BPE \citep{sennrich:2016:neural}.
We compare it to \gbs and demonstrate empirically that it is significantly faster, making constrained decoding with an arbitrary number of constraints feasible with GPU-based inference.
We also use the algorithm to study beam search interactions between model and metric scores, beam size, and pruning.

\section{Beam Search and Grid Beam Search}

Inference in statistical machine translation seeks to find the output sequence, $\hat{y}$, that maximizes the probability of a function parameterized by a model, $\theta$, and an input sequence, $x$:
$$
\hat{y} = \argmax_{y\in \cal{Y}} p_\theta(y \mid x)
$$
The space of possible translations, $\cal{Y}$, is the set of all sequences of words in the target language vocabulary, $V_T$.
It is impossible to explore this entire space.
Models decompose this problem into a sequence of time steps, $t$.
At each time step, the model produces a distribution over $V_T$.
The simplest approach to translation is therefore to run the steps of the decoder, choosing the most-probable token at each step, until either the end-of-sentence token, \eos, is generated, or some maximum output length is reached.
An alternative, which explores a slightly larger portion of the search space, is beam search.

In beam search \citep{lowerre:1976:harpy,sutskever:2014:sequence}, the decoder maintains a \emph{beam} of size $k$ containing a set of active hypotheses (Algorithm~\ref{algorithm:beam-search}).
At each time step $t$, the decoder model is used to produce a distribution over the target-language vocabulary, $V_T$, for each of these hypotheses.
This produces a large matrix of dimensions $k \times |V_T|$, that can be computed quickly with modern GPU hardware.
Conceptually, a (row,~column) entry $(i,j)$ in this matrix contains the state obtained from starting from the $i$th state in the beam and generating the target word corresponding to the $j$th word of $V_T$.
The beam for the next time step is filled by taking the states corresponding to the $k$-best items from this entire matrix and sorting them.


\begin{algorithm}[t]
  \caption{Beam search. \emph{Inputs:} max output length $N$, beam size $k$. \emph{Output:} highest-scoring hypothesis.}
  \label{algorithm:beam-search}
  \begin{algorithmic}[1]
  \Function{beam-search}{$N, k$}
  \State beam $\gets$ \Call{decoder-init}{$k$}
  \For{time step $t$ in 1..$N$}
  \State scores = \Call{decoder-step}{beam}
  \State beam $\gets$ \Call{kbest}{scores}\label{line:beam-search:kbest}
  \EndFor
  \State \Return beam[0]
  \EndFunction

  \Function{kbest}{scores}
  \State beam = \Call{argmax\_k}{$k$, scores}
  \State \Return beam
  \EndFunction
  \end{algorithmic}
\end{algorithm}


A principal difference between beam search for phrase-based and neural MT is that in NMT, there is no recombination: each hypothesis represents a complete history, back to the first word generated.
This makes it easy to record properties of the history of each hypothesis that were not possible with dynamic programming.
\citet{hokamp:2017:lexically} introduced an algorithm for forcing certain words to appear in the output called \emph{grid beam search} (\gbs).
This algorithm takes a set of constraints, which are words that must appear in the output, and ensures that hypotheses have met all these constraints before they can be considered to be completed.
For $C$ constraints, this is accomplished by maintaining $C+1$ separate beams or \emph{banks}, $B_0, B_1, \dots, B_C$, where $B_i$ groups together hypotheses that have generated (or \emph{met}) $i$ of the constraints.
Decoding proceeds as with standard beam decoding, but with the addition of bookkeeping that tracks the number of constraints met by each hypothesis, and ensures that new candidates are generated, such that each bank is filled at each time step.
When beam search is complete, the hypothesis returned is the highest-scoring one in bank $B_C$.
Conceptually, this can be thought of as adding an additional dimension to the beam, since we multiply out some base beam size $b$ by (one plus) the number of constraints.

We note two problems with \gbs:
\begin{itemize}
\item Decoding complexity is linear in the number of constraints:
  The effective beam size, $k \cdot (C+1)$, varies with the number of constraints.
\item
  It is impractical.
  The beam size changes for every sentence, whereas most decoders specify the beam size at model load time in order to optimize computation graphs, specially when running on GPUs.
  It also complicates beam search optimizations that increase throughput, such as batching.
\end{itemize}
Our extension, fast lexically-constrained decoding via dynamic beam allocation (\dba), addresses both of these issues.
Instead of maintaining $C+1$ beams, we maintain a single beam of size $k$, as with unconstrained decoding.
We then dynamically allocate the slots of this beam across the constraint banks at each time step.
There is still bookkeeping overhead, but this cost is constant in the number of constraints, instead of linear.
The result is a practical algorithm for incorporating arbitrary target-side constraints that fits within the standard beam-decoding paradigm.

\section{Dynamic Beam Allocation (\dba)}

Our algorithm (Algorithm~\ref{alg:dba}) is based on a small but important alteration to \gbs.
Instead of \emph{multiplying} the beam by the number of constraints, we \emph{divide}.
A fixed beam size is therefore provided to the decoder, just as in standard beam search.
As different sentences are processed with differing numbers of constraints, the beam is dynamically allocated to these different banks.
In fact, the allocation varies not just by sentence, but across time steps in processing each individual sentence.

We need to introduce some terminology.
A \emph{word constraint} provided to the decoder is a single token in the target language vocabulary.
A \emph{phrasal constraint} is a sequence of two or more contiguous tokens.
Phrasal constraints come into play when the user specifies a multi-word phrase directly (e.g., \emph{high-ranking member}), or when a word gets broken up by subword splitting (e.g., \emph{thou@\!@ ghtful}).
The total number of constraints is the sum of the number of tokens across all word and phrasal constraints.
It is easier for the decoder to place multiple sequential tokens in a phrasal constraint (where the permutation is fixed) compared to placing separate, independent constraints (see discussion at the end of \S\ref{section:validation}), but the algorithm does not distinguish them when counting.

\dba fits nicely within standard beam decoding; we simply replace the \verb|kbest| implementation from Algorithm~\ref{algorithm:beam-search} with one that involves a bit more bookkeeping.
Instead of selecting the top-$k$ items from the $k \times V_T$ scores matrix, the new algorithm must consider two important matters.
\begin{enumerate}
\item Generating a list of candidates (\S\ref{section:generating}).
  Whereas the baseline beam search simply takes the top-$k$ items from the scores matrix (a fast operation on a GPU), we now need to ensure that candidates progress through the set of provided constraints.
\item Allocating the beam across the constraint banks (\S\ref{section:allocating}).
  With a fixed-sized beam and an arbitrary number of constraints, we need to find an allocation strategy for dividing the beam across the constraint banks.
\end{enumerate}

\begin{figure*}[t]
  \begin{center}
    \includegraphics[width=158mm]{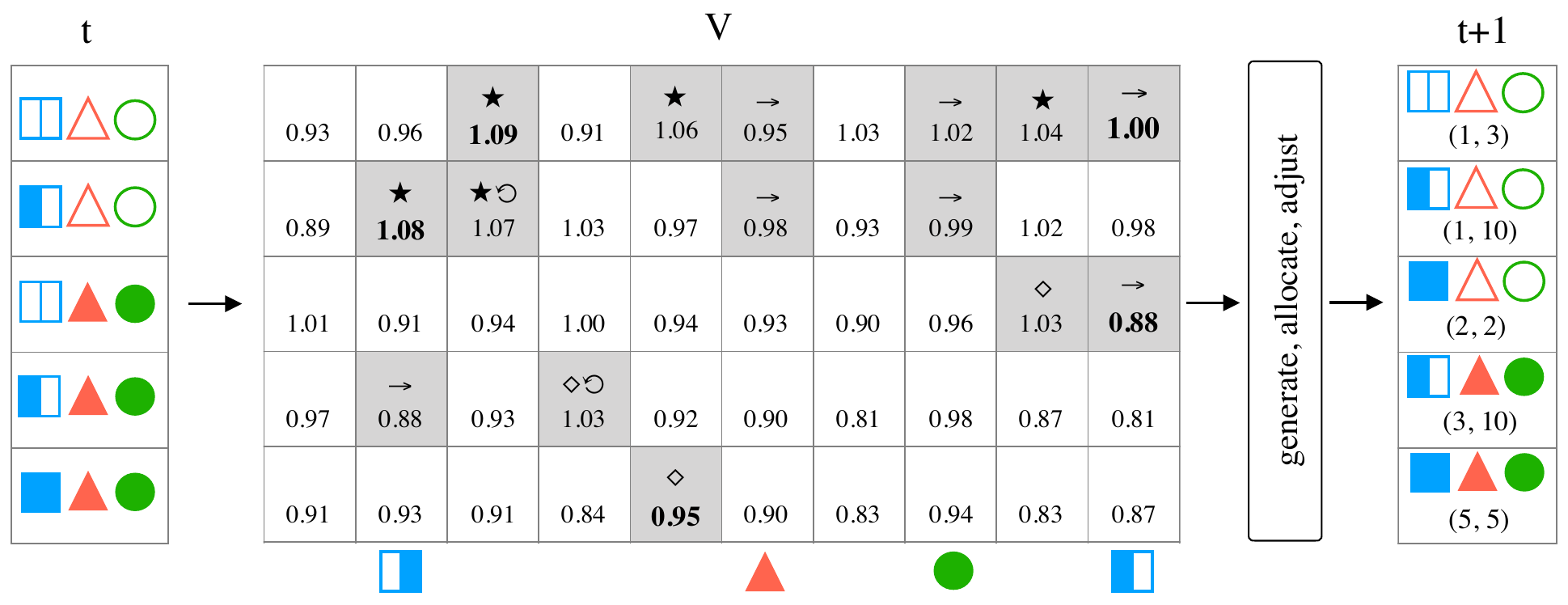}
  \end{center}
  \caption{A single step of the constrained decoder.
    Along the left is the beam ($k=5$) at time step $t$.
    The shapes in this beam represent constraints, both met (filled) and unmet (outlined).
    The blue square represents a phrasal constraint of length 2, which must be completed in order (left half, then right half).
    A step of the decoder produces a $k \times V_T$ matrix of scores.
    Each constraint corresponds to a single token in the vocabulary, and is marked along the bottom.
    Gray squares denote the set of candidates that are produced (\S\ref{section:generating}) from the $k$ best items (\blackstar), from extending each hypothesis with all unfilled constraints ($\rightarrow$), and from its single-best next token (\whitestar).
    Items that violate a phrasal constraint (\undo) require the phrasal constraint from that hypotheses to be unwound (set to unmet).
    From these fifteen candidates, the beam at time step $t+1$ is filled, according to the bank allocation strategy, which here assigns one slot in the beam to each bank.
    The final beam includes coordinates indicating the provenance of chosen items (which are also indicated in bold in the grid).}
  \label{figure:beam-search}
\end{figure*}

\begin{algorithm*}[t]
  \caption{$k$-best extraction with \dba. \emph{Inputs:} A $k \times |V_T|$ matrix of model states.}
  \label{alg:dba}
  \label{algorithm:dba}
  \begin{algorithmic}[1]
  \Function{kbest-\dba}{beam, scores}
  \State constraints $\gets$ [hyp.constraint for hyp in beam]
  \State candidates $\gets$ [$(i,j,$ constraints[$i$].add($j$)] for $i,j$ in \Call{argmax\_k}{$k$, scores} \Comment{Top overall $k$}
  \For{$1 \leq h \leq k$} \Comment{Go over current beam}
    \ForAll{$w \in V_T$ that are unmet constraints for beam[$h$]} \Comment{Expand new constraints}
      \State candidates.append( $(h,w,$ constraints[$h$].add($w$) ) )
    \EndFor
    \State $w$ = \Call{argmax}{scores[$h, :$]}
    \State candidates.append( ($h, w$, \, constraints[$h$].add($w$)) ) \Comment{Best single word}
  \EndFor
  \State selected $\gets$ \Call{allocate}{candidates, $k$}
  \State newBeam $\gets$ [candidates[$i$] for $i$ in selected]

  \State \Return newBeam
  \EndFunction
  \end{algorithmic}
\end{algorithm*}

\subsection{Generating the candidate set}
\label{section:generating}

We refer to Figure~\ref{figure:beam-search} for discussion of the algorithm.
The set of candidates for the beam at time step $t+1$ is generated from the hypotheses in the current beam at step $t$, which are sorted in decreasing order, with the highest-scoring hypothesis at position~1.
The \Call{decoder-step}{} function of beam search generates a matrix, $\mathit{scores}$, where each row~$r$ corresponds to a probability distribution over all target words, expanding the hypothesis in position~$r$ in the beam.
We build a set of candidates from the following items:
\begin{enumerate}
    \item The best $k$ tokens across all rows of $\mathit{scores}$ (i.e., normal top-$k$);
    \item for each hypothesis in the beam, all unmet constraints (to ensure progress through the constraints); and
    \item for each hypothesis in the beam, the single-best token (to ensure consideration of partially-completed hypotheses).
\end{enumerate}
Each of these candidates is denoted by its coordinates in \verb|scores|.
The result is a set of candidates which can be grouped into banks according to how many constraints they have met, and then sorted within those banks.
The new beam for timestep $t+1$ is then built from this list according to an allocation policy (next section).

For hypotheses partially through a phrasal constraint, special care must be taken.
If a phrasal constraint has been begun, but not finished, and a token is chosen that does not match the next word of the constraint, we must reset or ``unwind'' those tokens in this constraint that are marked as having been met.
This permits the decoder to abort the generation of a phrasal constraint, which is important in situations where a partial prefix of a phrasal constraint appears in the decoded sentence earlier than the entire phrase.

\subsection{Allocating the beam}
\label{section:allocating}

The task is to allocate a size-$k$ beam across $C+1$ constraint banks, where $C$ may be greater than $k$.
We use the term \emph{bank} to denote the portion of the beam reserved for items having met the same number of constraints (including one bank for hypotheses with zero constraints met).
We use a simple allocation strategy, setting each bin size to ${\lfloor}k/C{\rfloor}$, irrespective of the timestep.
Any remaining slots are assigned to the ``topmost'' or maximally constrained bank, $C$.

\begin{figure}[t]
  \begin{center}
    \includegraphics[width=75mm]{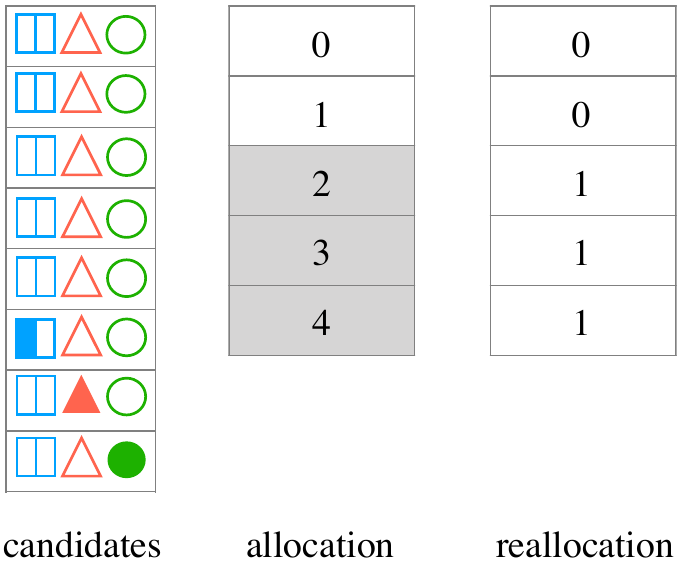}
  \end{center}
  \caption{Beam reallocation for $k=5$ with 4 constraints at timestep $t$.
    There are eight candidates, each having met only 0 or 1 constraint.
    The allocation policy gives one slot of the beam to each bank.
    However, there are no candidates for banks 2--4 (greyed), so their slots are redistributed to banks 0 and 1.}
  \label{figure:adjustment}
\end{figure}

This may at first appear wasteful.
For example, space allocated at timestep 1 to a bank representing candidates having met more than one constraint cannot be used, and similarly, for later timesteps, it seems wasteful to allocate space to bank 1.
Additionally, if the number of candidates in a bank is smaller than the allocation for that bank, the beam is in danger of being underfilled.
These problems are mitigated by bank adjustment (Figure~\ref{figure:adjustment}).
We provide here only a sketch of this procedure.
An overfilled bank is one that has been allocated more slots than it has candidates to fill.
Each such overfilled bank, in turn, gives its extra allotments to banks that have more candidates than slots, looking first to its immediate neighbors, and moving outward until it has distributed all of its extra slots.
In this way, the beam is filled, up to the minimum of the beam size or the number of candidates.

\subsection{Finishing}
\label{section:finishing}

Hypotheses are not allowed to generate the end-of-sentence token, \eos, unless they have met all of their constraints.
When beam search is finished, the highest-scoring completed item is returned.

\section{Experimental Setup}
\label{section:setup}

Our experiments were done using \sockeye \citep{hieber:2017:sockeye}.
We used an English--German model trained on the complete WMT'17 training corpora \citep{bojar:2017:findings}, which we preprocessed with the Moses tokenizer (preserving case) and with a joint byte-pair-encoded vocabulary with 32k merge operations \citep{sennrich:2016:neural}.
The model was a 4 layer RNN with attention.
We trained using the Adam optimizer with a batch size of 80 until cross-entropy on the development data (newstest2016) stopped increasing for 10 consecutive iterations.

For decoding, we normalize completed hypotheses (those that have generated \eos), dividing the cumulative sentence score by the number of words.
Unless otherwise noted, we apply threshold pruning to the beam, removing hypotheses whose log probability is not within 20 compared to the best completed hypothesis.
This pruning is applied to all hypotheses, whether they are complete or not.
(We explore the importance of this pruning in \S\ref{section:pruning}).
Decoding stops when either all hypotheses still on the beam are completed or the maximum length, $N$, is reached.
All experiments were run on a single a Volta P100 GPU.
No ensembling or batching were used.

For experiments, we used the newstest2014 English--German test set (the developer version, with 2,737 sentences).
All BLEU scores are computed on detokenized output using \sacrebleu \cite{post:2018:call},\footnote{The signature is BLEU+case.mixed+lang.en-de+numrefs.1+smooth.exp+test.wmt14+tok.13a+v.1.2.6} and are thus directly comparable to scores reported in the WMT evaluations.

\section{Validation Experiment}
\label{section:validation}

We center our exploration of \dba by experimenting with constraints randomly selected from the references.
We extract five sets of constraints: from one to four randomly selected words from the reference (\verb|rand1| to \verb|rand4|), and a randomly selected four-word phrase (\verb|phr4|).
We then apply BPE to these sets, which often yields a much larger number of token constraints.
Statistics about these extracted phrases can be found in Table~\ref{table:phrases}.

\begin{table}[t]
  \centering\centering\resizebox{\columnwidth}{!}{
    \begin{tabular}{l|rrrrr}
      num & rand1 & rand2 & rand3 & rand4 & phr4 \\
      \hline\hline
      1 & 2,182 & 0 & 0 & 0 & 0 \\
      2 & 548 & 3,430 & 0 & 0 & 0 \\
      3 & 516 & 1,488 & 4,074 & 0 & 0 \\
      4 & 272 & 1,128 & 2,316 & 4,492 & 4,388 \\
      5 & 150 & 765 & 1,860 & 3,275 & 2,890 \\
      6 & 30 & 306 & 1,218 & 2,520 & 2,646 \\
      7 & 42 & 133 & 805 & 1,736 & 1,967 \\
      8 & 0 & 112 & 488 & 1,096 & 1,280 \\
      9 & 0 & 36 & 171 & 702 & 720 \\
      10 & 0 & 10 & 140 & 400 & 430 \\
      11+ & 0 & 22 & 189 & 417 & 575 \\
      \hline
      total &  3,726 &  7,477 & 11,205 & 14,885 & 14,926 \\
      mean  &   1.36 &   2.73 &   4.09 &   5.43 &   5.45 \\
    \end{tabular}
  }
  \caption{Histogram of the number of token constraints for some constraint sets after applying BPE (model trained with 32k merge operations).
    \emph{mean} denotes the mean number of constraints per sentence in the 2,737-sentence test set.}
  \label{table:phrases}
\end{table}

We simulate the GBS baseline within our framework.
After applying BPE, We group together translations with the same number of constraints, $C$, and then translate them as a group, with the beam set for that group set to $b(C+1)$, where $b$ is the ``base beam'' parameter.
We use $b=10$ as reported in Hokamp et al., but also try smaller values of $b=5$ and 1.
Finally, we disable beam adjustment (\S\ref{section:allocating}), so that the space allocated to each constraint bank does not change.

\begin{figure}[t]
  \begin{center}
    \includegraphics[width=75mm]{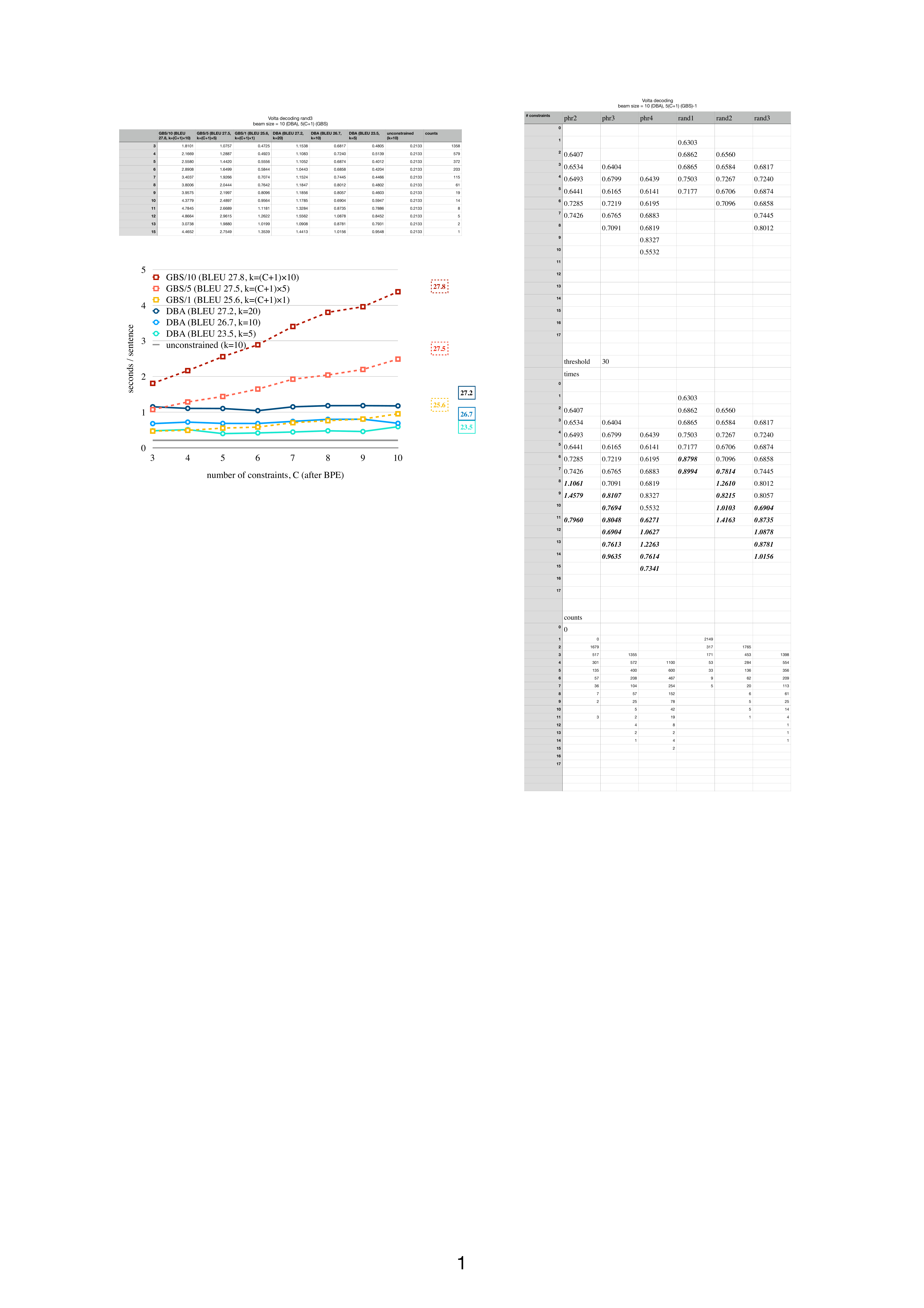}
    \caption{Running time (seconds / sentence, lower is better) as a function of the number of constraints, $C$ (after applying BPE) on the \texttt{rand3} dataset.
      The unconstrained baselines have BLEU scores of 22.3, 22.3, and 22.1 for $k=5,10$, and 20, respectively.}
    \label{figure:speeds}
  \end{center}
\end{figure}

Table~\ref{figure:speeds} compares speeds and BLEU scores (in the legend) as a function of the number of post-BPE constraints for the \verb|rand3| dataset.
We plot all points for which there were at least 10 sentences.
The times are decoding only, and exclude model loading and other setup.
The linear trend in $C$ is clear for GBS, as is the constant trend for DBA.
In terms of absolute runtimes, DBA improves considerably over GBS, whose beam sizes quickly become quite large with a non-unit base beam size.
On the Tesla V100 GPU, DBA ($k=10$) takes about 0.6 seconds/sentence, regardless of the number of constraints.\footnote{On a K80, it is about 1.4 seconds / sentence}
This is about 3x slower than unconstrained decoding.

It is difficult to compare these algorithms exactly because of \gbs's variable beam size.
An important comparison is that between DBA ($k=10$) and GBS/1 (base beam of 1).
A beam of $k=10$ is a common setting for decoding in general, and GBS/1 has a beam size of $k\ge10$ for $C\ge9$.
At this setting, DBA finds better translations (BLEU 26.7 vs.\ 25.6) with the same runtime and with a fixed, instead of variable-sized, beam.

\begin{figure}[t]
  \begin{center}
    \includegraphics[width=75mm]{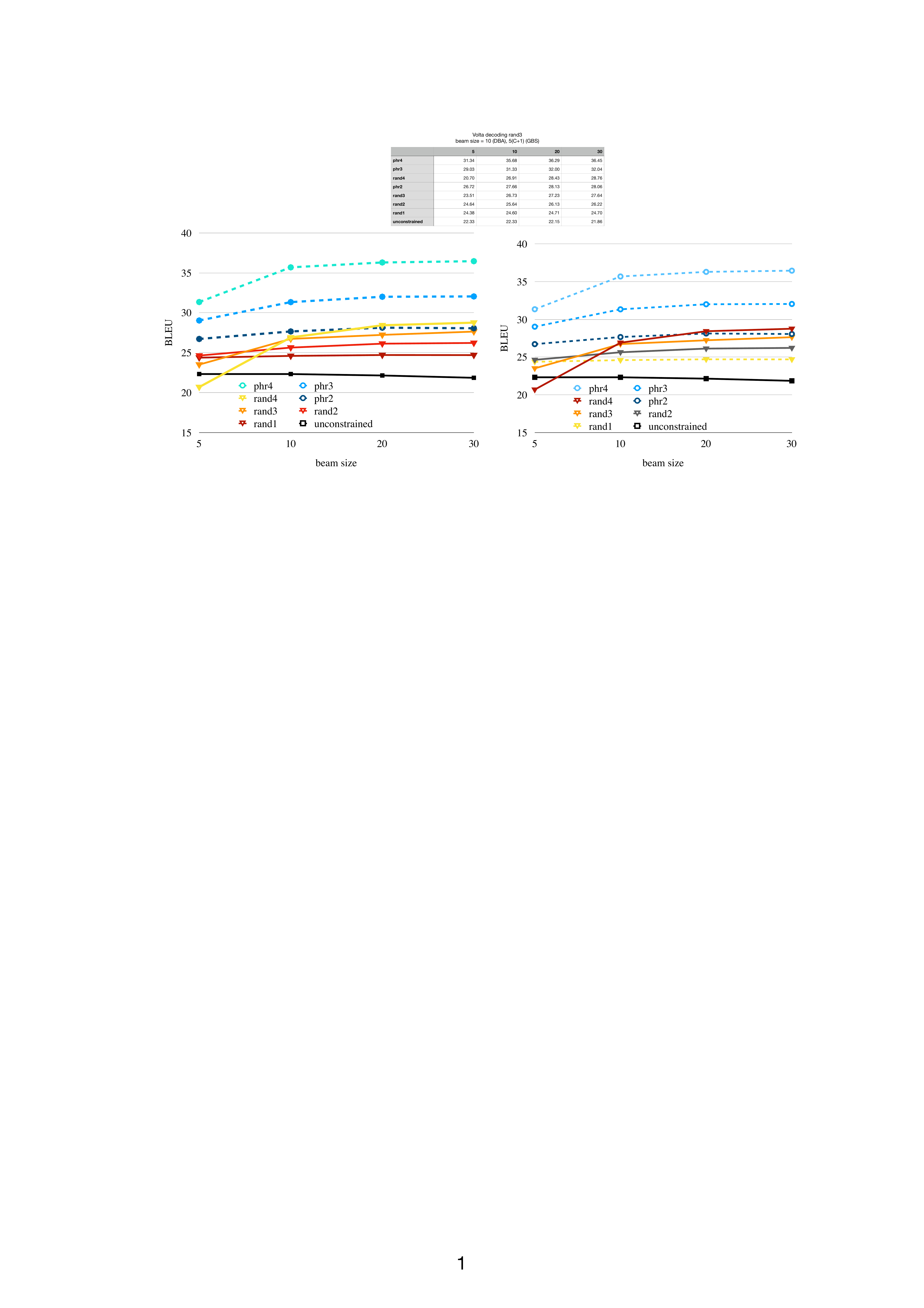}
  \end{center}
  \caption{BLEU score as a function of beam size under DBA.
    All constraint sets improve as the beam gets larger (recall that the actual number of constraints increases after BPE and varies by sentence).
    \texttt{rand4} performs under the unconstrained baseline if the beam is too low.}
  \label{figure:beam-size}
\end{figure}

We note that the bank adjustment correction of the DBA algorithm allows it to work when $C>=k$.
The DBA ($k=5$) plot demonstrates this, while still finding a way to increase the BLEU score over GBS (23.5 vs.\ 22.3).
However, while possible, low $k$ relative to $C$ reduces the observed improvement considerably.
Looking at Figure~\ref{figure:beam-size} across different constraint sets, we can get a better feel for this relationship.
\dba is still always able to meet the constraints even with a beam size of 5, but the quality suffers.
This should not be too surprising; correctly placing independent constraints is at least as hard as finding their correct permutation, which is exponential in the number of independent constraints.
But it is remarkable that the only failure to beat the baseline in terms of BLEU is when the algorithm is tasked with placing four random constraints (before BPE) with a beam size of 5.
In contrast, \dba never has any trouble placing phrasal constraints (dashed lines).

\section{Analysis}

\subsection{Placement}

\begin{figure*}
    \resizebox{\textwidth}{!}{
    \begin{tabular}{lrp{110mm}}
      constraint & score & output \\
      \hline\hline
      source & &
      \emph{Einer soll ein hochrangiges Mitglied aus Berlin gewesen sein .}
      \\
      \hline
      no constraints & -0.217 &
	    One should have been a high-ranking member from Berlin .
      \\
      \emph{is said to} & -0.551 &
      One \underline{is said to} have been a high-ranking member from Berlin .
      \\
      \emph{of them} & -0.577 &
	    One \underline{of them} was to be a high-ranking member from Berlin .
      \\
      \emph{participant} & -0.766 &
      One should have been a high-ranking \underline{participant} from Berlin .
      \\
      \emph{is thought to} & -0.792 &
	    One \underline{is thought to} have been a high-ranking member from Berlin .
      \\
      \emph{considered} & -0.967 &
      One is \underline{considered} to have been a high-ranking member from Berlin .
      \\
      \emph{Hamburg} & -1.165 & One should have been a high-ranking member from \underline{Hamburg} . \\
      \emph{powerful} & -1.360 &
	    One is to have been a \underline{powerful} member from Berlin .
      \\
      \emph{powerful}, \emph{is said to} & -1.496 &
      One \underline{is said to} have been a \underline{powerful} member from Berlin .
      \\
      \emph{powerful}, \emph{is said to}, \emph{participant} & -1.988 &
      One \underline{is said to} have been a \underline{powerful} \underline{participant} from Berlin .
      \\
      \emph{weak} & -1.431 &
      One \underline{weak} point was to have been a high-ranking member from Berlin .
      \\
      \hline
      reference & &
      \emph{One is said to have been a high-ranking member from Berlin.}
      \\
    \end{tabular}
    }
    \caption{Example demonstrating the correct placement of manually chosen constraints (beam size 10).
      The unnatural placement of the constraint \emph{weak} demonstrates what the model does when forced to include a word that is not a semantic fit.}
  \label{figure:placement-examples}
\end{figure*}

It's possible that the BLEU gains result from a boost in n-gram counts due to the mere presence of the reference constraints in the output, as opposed to their correct placement.
This appears not to be the case.
Experience examining the outputs shows its uncanny ability to sensibly place constrained words and phrases.
Figure~\ref{figure:placement-examples} contains some examples from translating a German sentence into English, manually identifying interesting phrases in the target, choosing paraphrases of those words, and then decoding with them as constraints.
Note that the word weak, which doesn't fit in the semantics of the reference, is placed haphazardly.

\begin{figure}[t]
  \begin{center}
    \includegraphics[width=75mm]{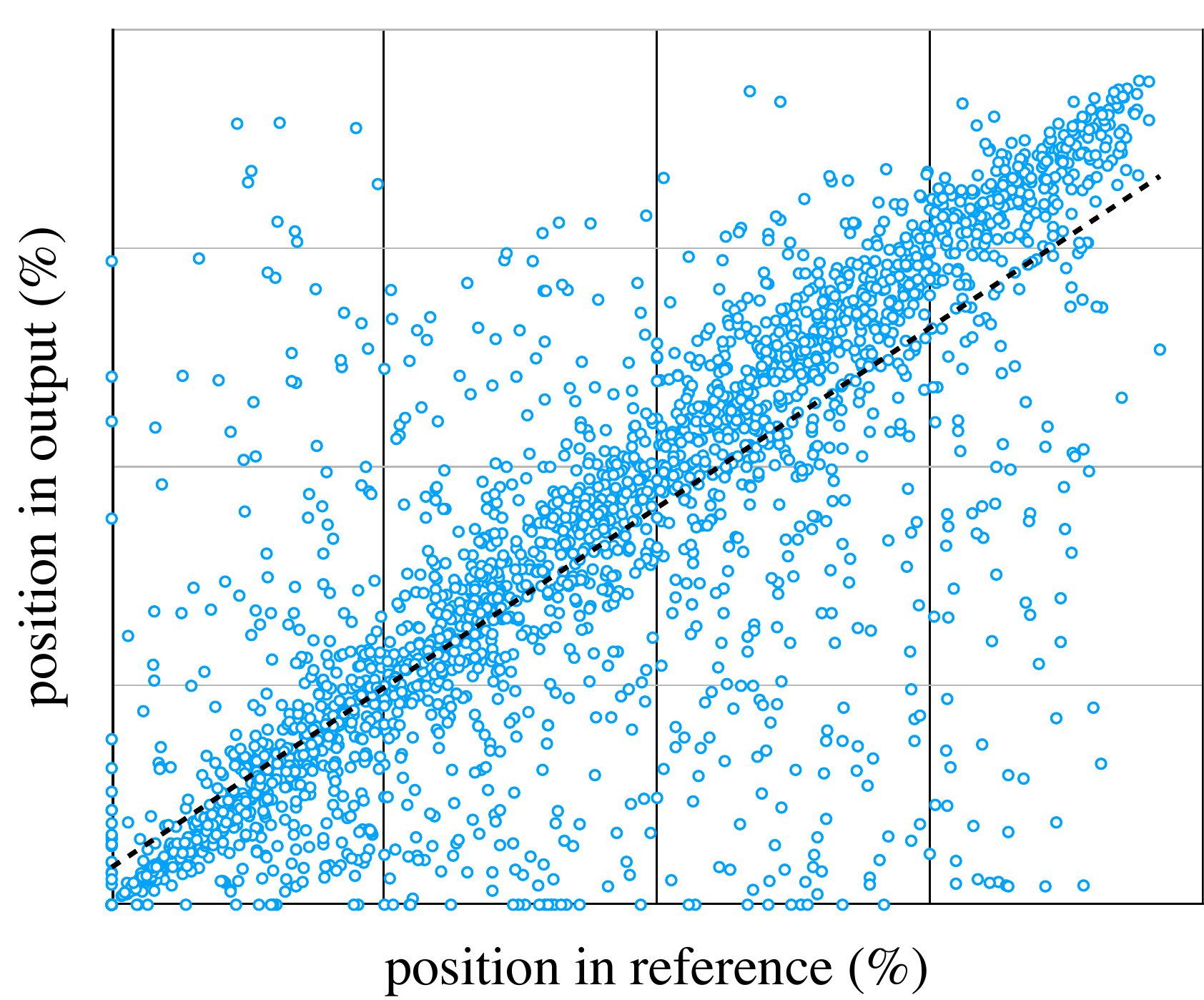}
  \end{center}
  \caption{Location of the first word of each constraint from \texttt{phr3} in the reference versus the constrained output (Pearson's $r=0.82$).
    \dba correctly places its constraints, even though no source word or alignment information is provided.}
  \label{figure:placement}
\end{figure}

We also confirm this correct placement quantitatively by comparing the location of the first word of each constraint in (a) the reference and (b) the output of the constrained decoder, represented as a percentage of the respective sentence lengths (Figure~\ref{figure:placement}).
We would not expect these numbers to be perfectly matched, but the strong correlation is pretty apparent (Pearson's $r=0.82$).
Together, Figures~\ref{figure:placement-examples} and \ref{figure:placement} provide confidence that \dba is intelligently placing the constraints.

\subsection{Reference Aversion}

The inference procedure in \sockeye maximizes the length-normalized version of the sentence's log probability.
While there is no explicit training towards the metric, BLEU, modeling in machine translation assumes that better model scores correlate with better BLEU scores.
However, a general repeated observation from the NMT literature is the disconnect between model score and BLEU score.
For example, work has shown that opening up the beam to let the decoder find better hypotheses results in lower BLEU score \cite{koehn:2017:six}, even as the model score rises.
The phenomenon is not well understood, but it seems that NMT models have learned to travel a path straight towards their goal; as soon as they get off this path, they get lost, and can no longer function \citep{ott:2018:analyzing}.

\begin{figure}[t]
  \begin{center}
    \includegraphics[width=75mm]{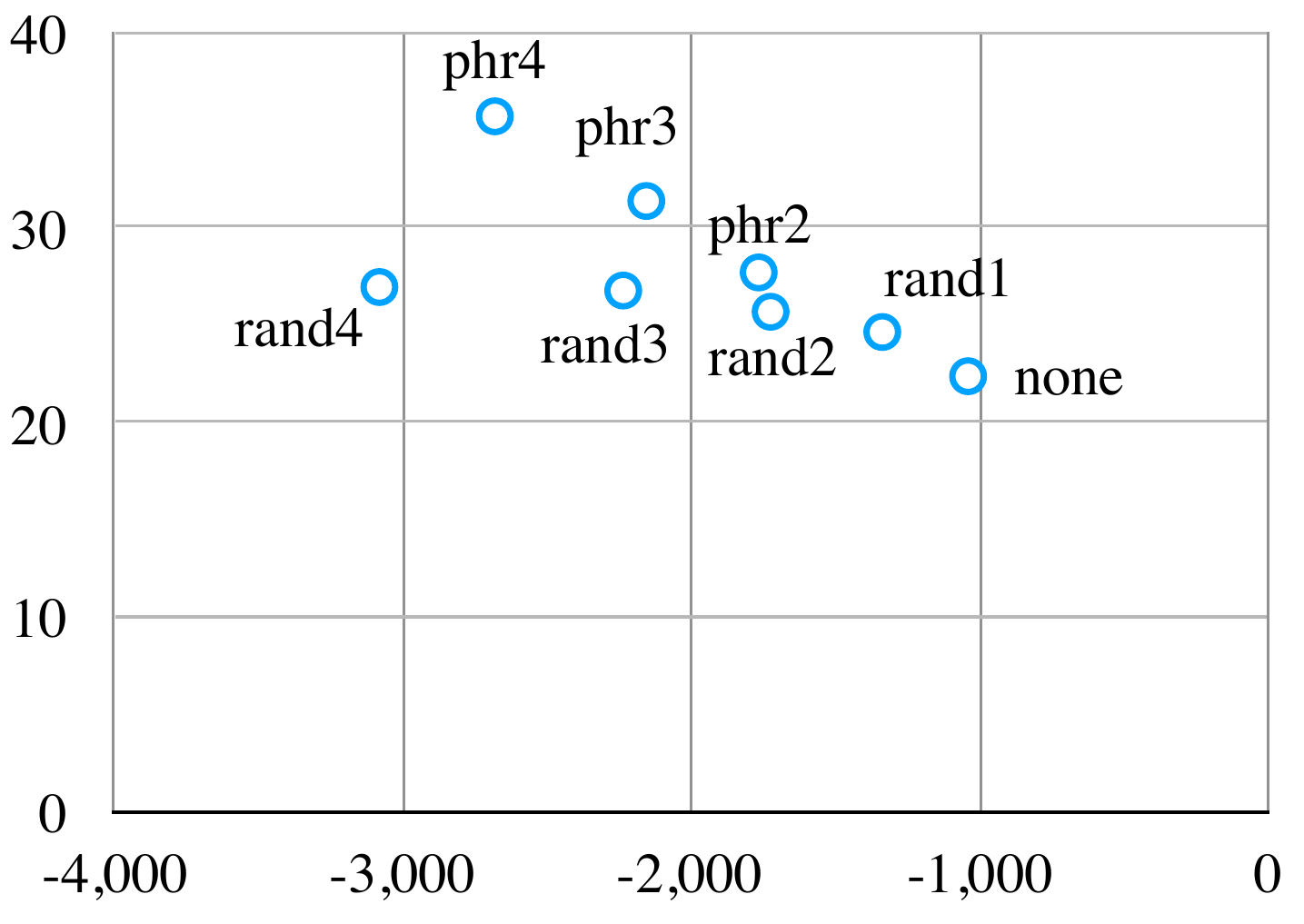}
    \caption{BLEU score as a function of model score (summed over the corpus).
      The reference model score is -4,396.}
    \label{figure:reference-aversion}
  \end{center}
\end{figure}

Another way to look at this problem is to ask what the neural model thinks of the references.
Scoring against complete references is easy with NMT \citep{sennrich:2017:grammatical}, but lexically-constrained decoding allows us to investigate this in finer-grained detail by including just portions of the references.
We observe that forcing the decoder to include even a single word from the reference imposes a cost in model score that is inversely correlated with BLEU score, and that this grows with the number of constraints that are added (Figure~\ref{figure:reference-aversion}).
The NMT system seems quite averse to the references, even in small pieces, and even while it improves the BLEU score.
At the same time, the hypotheses it finds in this reduced space are still good, and become better as the beam is enlargened (Figure~\ref{figure:beam-size}).
This provides a complementary finding to that of \citet{koehn:2017:six}: in that setting, higher model scores found by a larger beam produce lower BLEU scores; here, lower model scores are associated with significantly higher BLEU scores.

\begin{figure*}[t]
  \centering\resizebox{\textwidth}{!}{
    \begin{tabular}{rrrrrrrrrrrrrrrrrrrrrrrrrrrrr}
      & 1 & 2 & 3 & 4 & 5 & 6 & 7 & 8 & 9 & 10 & 11 & 12 & 13 & 14 & \dots & 41 \\
      \hline
      -0.51 & \sos & Er & und & Kerr & lieben & einander & noch & immer & , & betonte & die & 36-J\"{a}hrige & . & \eos \\
      -0.52 & \sos & Er & und & Kerr & lieben & einander & noch & immer & , & betonte & der & 36-J\"{a}hrige & . & \eos \\
      -0.56 & \sos & Er & und & Kerr & lieben & einander & noch & immer & , & betonte & die & 36-j\"{a}hrige & . & \eos \\
      -0.57 & \sos & Er & und & Kerr & lieben & einander & noch & immer & , & betonte & den & 36-j\"{a}hrigen & . & \eos \\
      -25.11 & \sos & Er & und & Kerr & lieben & sich & weiterhin & einander & , & betonte & die & 36-J\"{a}hrige & . & \&\#160; & \dots & \&\#160; \\
      -27.92 & \sos & Er & und & Kerr & lieben & sich & weiterhin & einander & , & betonte & die & 36-J\"{a}hrige & . & \&\#160; & \dots & \&\#160; \\
    \end{tabular}
}
  \caption{The sentence \emph{He and Kerr still love each other , emphasised the 36-year-old .} translated with the constraint \emph{noch immer , betonte} (BPE removed for readability).
    The first column is the log probability, which is normalized only for finished hypotheses.
    The decoder completes a few hypotheses well before the maximum timestep, but then fills the lower beam with garbage until forced to stop.}
  \label{figure:garbage}
\end{figure*}

\subsection{Effects of Pruning}
\label{section:pruning}

\begin{table}[t]
  \begin{center}
      \begin{tabular}{l|rrrrrr}
 & 0 & 3 & 5 & 10 & 20 & 30 \\
\hline
\hline
\small none & 24.4 & 24.5 & 24.5 & 24.4 & 24.5 & 24.4 \\
\hline
\small rand1 & 25.2 & 25.1 & 25.2 & 25.6 & 25.5 & 25.3 \\
\small rand2 & 26.0 & 25.3 & 25.6 & 26.1 & 26.7 & 26.4 \\
\small rand3 & 26.5 & 24.7 & 24.9 & 25.7 & 26.9 & 27.2 \\
\small rand4 & 26.2 & 23.7 & 23.9 & 24.6 & 26.0 & 26.9 \\
\small phr4 & 35.1 & 33.5 & 33.5 & 34.0 & 35.0 & 35.9 \\
    \end{tabular}
    \caption{BLEU scores varying the pruning setting (beam size 10).
      Runtimes for unpruned systems (column 0) are nearly twice those of the other columns.
      But it is only at large thresholds that BLEU scores are higher than the unpruned setting.}
    \label{table:pruning}
  \end{center}
\end{table}

In the results reported above, we used a pruning threshold of 20, meaning that any hypothesis whose log probability is not within 20 of the best completed hypothesis is removed from the beam.
This pruning threshold is far greater than those explored in other papers; for example, \citet{wu:2016:googles} use 3.
However, we observed two things: first, without pruning, running time for constrained decoding is nearly doubled.
This increased runtime applies to both \dba and \gbs in Figure~\ref{figure:speeds}.
Second, low pruning thresholds are harmful to BLEU scores (Table~\ref{table:pruning}).
It is only once the thresholds reach 20 that the algorithm is able to find better BLEU scores compared to the unpruned baseline (column 0).

\subsection{Garbage Generation}
\label{section:garbage}

Why is the algorithm so slow without pruning?
One might suspect that the outputs are longer, but mean output length with all constraint sets is roughly the same.
The reason turns out to be that the the decoder never quits before the maximum timestep, $N$.
\sockeye's stopping criterium is to wait until all hypotheses on the beam are finished.
Without pruning, the decoder generates a finished hypotheses, but continues on until the maximum timestep $N$, populating the rest of the beam with low-cost garbage.
An example can be found in Figure~\ref{figure:garbage}.
This may be an example of the well-attested phenomenon where NMT systems become unhinged from the source sentence, switching into ``language model'' mode and generating high-probable output with no end.
But strangely, this doesn't seem to affect the best hypotheses, but only the rest of the beam.
This seems to be more evidence of reference aversion, where the decoder, having been forced into a place it doesn't like, does not know how to generate good competing hypotheses.

An alternative to pruning is \emph{early stopping}, which is to stop when the first complete hypothesis is generated.
In our experiments, while this did fix the problem of increasing runtimes, the BLEU scores were lower.

\section{Related Work}

\citet{hokamp:2017:lexically} was novel in that it allowed the specification of arbitrary target-side words as hard constraints, implemented entirely as a restructuring of beam search.
A related approach was that of \citet{anderson:2017:guided}, who similarly extended beam search with a finite state machine whose states marked completed subsets of the set of constraints, at an exponential cost in the number of constraints.
Both of the approaches make use of target-side constraints without any reference to the source sentence.
A related line of work is to attempt to force the use of particular translations of source words and phrases, which allows the incorporation of external resources such as glossaries or translation memories.
\citet{chatterjee:2017:guiding} propose a variant on beam search that interleaves decoder steps with attention-guided checks to ensure that glossary annotations on the source sentence are generated by the decoder.
Their implementation also permits discontinuous phrases and translation options (these capabilities are not available in \dba, but could be added).
A softer approach treats external resources as suggestions that are allowed to influence, but not necessarily determine, the choice of words \cite{arthur:2016:incorporating} and also phrases \citep{tang:2016:neural}.

Lexically-constrained decoding also generalizes prefix decoding \cite{knowles:2016:neural,wuebker:2016:models}, since the \sos symbol can easily be included as the first word of a constraint.

Finally, another approach which shares the hard-decision made by lexically constrained decoding is the \emph{placeholder approach} \cite{crego:2016:systrans}, wherein identifiable elements in the input are transformed to masks during preprocessing, and then replaced with their original source-language strings during postprocessing.

\section{Summary}

Neural machine translation removes many of the knobs from phrase-based MT that provided fine-grained control over system output.
Lexically-constrained decoding restores one of these tools, providing a powerful and interesting way to influence NMT output.
It requires only the specification of the target-side constraints; without any source word or alignment information, it correctly places the constraints.
Although we have only tested it here with RNNs, the code works without modification with other architectures generate target-side words one-by-one, such as the Transformer \cite{vaswani:2017:attention}.

This paper has introduced a fast and practical solution.
Building on previous approaches, constrained decoding with \dba does away with linear and exponential complexity (in the number of constraints), imposing only a constant overhead.
On a Volta GPU, lexically-constrained decoding with \dba is practical, requiring about 0.6 seconds per sentence on average even with 10+ constraints, well within the realm of feasibility even for applications with strict lattency requirements, like post-editing tasks.
We imagine that there are further optimizations in reach that could improve this even further.

\paragraph{Acknowledgments}

We thank Felix Hieber for valuable discussions.

\bibliography{everything}
\bibliographystyle{acl_natbib}

\appendix

\section{Appendix: Failed Experiments}

The main contribution of this paper is a fast, practical algorithmic improvement to lexically-constrained decoding.
While we did not attempt to corroborate the experiments in interactive translation and domain adaptation experiments reported in \citep{hokamp:2017:lexically}, the gains discovered there only become more salient with this faster algorithm.
We did try to apply lexical constraints in a few other settings, but without success.
In the spirit of open scientific inquiry and reporting, we provide here a brief report on these experiments.

\subsection{Automatic Constraint Selection}

Our validation experiments (\S\ref{section:validation}) demonstrate the large potential gains in BLEU score when including random phrases from the reference.
Even including just a single random word from the reference increased BLEU score by a point; another point was gained from including two random words, and a four-word phrase yielded 10+ point gains.
Only about 18\% of these random unigrams were present in the unconstrained output (less for longer n-grams).
This raises the question of whether we can automatically identify words that are likely to be in the reference and include them as constraints, in order to improve translation quality.

In order to do that, we first extracted a phrase table using Moses and filtered it with the significance testing approach proposed by \citet{johnson:2007:improving} in order to keep only high quality phrases.
We then selected the best phrase for each input sentence according to different criteria (longest phrase, higher significance, highest probability, combination of those).
Unfortunately, adding such phrases as constraints when translating WMT or IWSLT data did not help.

\subsection{Name Entity Translation}

One topic that has received attention in the literature is the tendency of NMT systems to do poorly with rare words, and in particular, named entities (e.g., \citet{arthur:2016:incorporating}).
BPE helps address this by breaking down words into pieces and allowing all words to be represented in the decoder's vocabulary.
But even with BPE, many times the correct translation does not follow any pattern, even at the subword level.
This is specially true for named entities; e.g.\ ``Aachen'' in German is translated as ``Aquisgr\'an'' in Spanish or ``Aix-la-Chapelle'' in French, which bear little resemblance to the original form except for the starting letter.
Named entities~(NEs) also have the advantage that in several languages they are not inflected; therefore a simple lookup in a dictionary, if available, should produce the correct translation.

To the best of our knowledge, there is no publicly available parallel corpus of named entities.
In order to create one, we downloaded the OpenSubtitles database \cite{lison:2016:opensubtitles2016} for German and English and applied a simple method for extracting named entity correspondences.
We first tagged the source and target sides with the Stanford NER system \citep{manning:2014:stanford}.
We then selected a subset of the tags that were produced by both systems (``Person'', ``Location'' and ``Organization'') and selected those sentences where they appeared only once for each language.
From those we extracted the corresponding NEs, selecting the most frequent target side at the corpus level as the translation of a given source NE.

Given such a dictionary, we can add the translation of a NE found in new sentences to translate as decoding constraints.
It didn't help.
Manual inspection showed that the dictionary extracted with this simple method was still too noisy.
We think that a manual, high-quality dictionary may provide a way to produce improvements.


\end{document}